\documentclass[journal]{IEEEtran}

\usepackage[colorlinks=false]{hyperref}
\usepackage{amsmath,url, amssymb}
\usepackage{mathrsfs}
\usepackage[pdftex]{graphicx}
\usepackage[caption=false]{subfig}
\usepackage{balance}
\usepackage{enumitem, kantlipsum}
\usepackage[mathscr]{euscript}
 \let\mathscr\relax
\usepackage[scr]{rsfso}
\usepackage{textcomp}
\usepackage{multirow, bm}
\usepackage{makecell}
\usepackage{arydshln}
\usepackage{xcolor}
\usepackage{afterpage}
\usepackage{tikz}
\usepackage{amsfonts}
\usepackage{tabularx}
\usepackage{float}
\usepackage{xcolor}
\usepackage{diagbox}
\newcolumntype{?}{!{\vrule width 2\arrayrulewidth}}
\usepackage{comment}
\def\I{{\mathbf I}}
\def\K{{\mathbf K}}
\def\W{{\mathbf W}}
\def\P{{\mathbf P}}

\def\pce{{\textrm{PCE}}}
\def\cs{{\mathrm{C}_s}}
\def\Dtv{{\mathcal{D}_\text{T}}}

\def\Acs{\mathrm{A}_{\mathsf{cs}}}
\def\Aos{\mathrm{AUC}_{\mathsf{os}}}
\def\auc{\mathrm{AUC}}

\begin{document}

\title{CNN-based fast source device identification}

\author{Sara~Mandelli,
        Davide Cozzolino,~\IEEEmembership{Member,~IEEE,}
        Paolo~Bestagini,~\IEEEmembership{Member,~IEEE,}
        Luisa~Verdoliva,~\IEEEmembership{Senior~Member,~IEEE,}
        and~Stefano~Tubaro,~\IEEEmembership{Senior~Member,~IEEE}
\thanks{S. Mandelli, P. Bestagini and S. Tubaro are with the Dipartimento di Elettronica, Informazione e Bioingegneria, Politecnico di Milano, Milano 20133, Italy (e-mail: name.surname@polimi.it).}
\thanks{D. Cozzolino is with the Dipartimento di Ingegneria Elettrica e Tecnologie dell'Informazione, University Federico II of Naples, 80125 Naples, Italy (e-mail: davide.cozzolino@unina.it).}
\thanks{L. Verdoliva is with the Dipartimento di Ingegneria Industriale, University Federico II of Naples, 80125 Naples, Italy (e-mail: verdoliv@unina.it).}
\thanks{This material is based on research sponsored by DARPA and AFRL under agreement number FA8750-16-2-0173. The U.S. Government is authorized to reproduce and distribute reprints for Governmental purposes notwithstanding any copyright notation thereon. The views and conclusions contained herein are those of the authors and should not be interpreted as necessarily representing the official policies or endorsements, either expressed or implied, of DARPA and AFRL or the U.S. Government. This work was supported by the PREMIER project, funded by the Italian Ministry of Education, University, and Research within the PRIN 2017 program. Hardware support was generously provided by the NVIDIA Corporation.}
}

\maketitle

\begin{abstract}
Source identification is an important topic in image forensics, since it allows to trace back the origin of an image.
This represents a precious information to claim intellectual property but also to reveal the authors of illicit materials.
In this paper we address the problem of device identification based on sensor noise and propose a fast and accurate solution using convolutional neural networks (CNNs).
Specifically, we propose a 2-channel-based CNN that learns a way of comparing camera fingerprint and image noise at patch level.
The proposed solution turns out to be much faster than the conventional approach and to ensure an increased accuracy.
This makes the approach particularly suitable in scenarios where large databases of images are analyzed, like over social networks.
In this vein, since images uploaded on social media usually undergo at least two compression stages, we include investigations on double JPEG compressed images, always reporting higher accuracy than standard approaches.

\end{abstract}


\section{Introduction}
\label{sec:introduction}

\IEEEPARstart{I}{n}
this paper, we tackle the problem of image source device identification. 
State-of-the-art solutions rely on photo response non-uniformity (PRNU) \cite{Lukas2006}, 
a characteristic noise trace left by the camera sensor on all acquired images.
PRNU is caused by imperfections in the sensor manufacturing process
and represents a unique pattern noise associated to a certain device.
It can be used both to trace the exact origin of an image and to establish its integrity \cite{Chen2008, Chierchia2014}.

The classic pipeline is based on the availability of a number of images coming from the same device in order to carry out a reliable PRNU estimation.
Then, the image under test can be compared with the sensor fingerprint through denoising and peak to correlation energy (PCE) computation \cite{Chen2007}.
However, computing the PCE between a query image noise and a fingerprint might be computationally expensive, and may become a major bottleneck when source identification requires scanning huge databases of device fingerprints \cite{Lin2017, Marra2017, Phan2018}.
With the increasing resolution of digital sensors, the search time may become prohibitive even for a moderate fingerprint database. 

To manage large image databases some methods look for compressed or alternative representations of the sensor noise.
One of the first methods introduces the concept of fingerprint digest: for each query fingerprint, only pixels with the largest magnitude are extracted \cite{Goljan2010}.
Instead, in \cite{Bayram2012} the idea is to reduce the resolution of PRNU values by quantizing them into a single bit, while in \cite{Bayram2015} a composite fingerprint is introduced.
In \cite{Valsesia2015} compression is obtained using random projections followed by binary quantization of the projected values, and in \cite{Bondi2019} this technique is investigated for JPEG images.
Alternatively, in \cite{Taspinar2017} a tree-like search strategy is designed to split PRNU matching problem into a series of sub-problems.

We consider an alternative approach to compare PRNUs and test images by leveraging convolutional neural networks (CNNs) to achieve good performance on small image patches.
Notice that relying on small patches makes PRNU-based forgery localization possible.
Moreover, it enables analyses on social networks, where images are often available only in cropped formats.
CNNs have been already successfully applied for camera model identification
\cite{Tuama2016, Bondi2017}, for tracing the origin of an image in a social network \cite{Caldelli2018} or for PRNU anonymization \cite{Bonettini2018}. However, less attention has been devoted to PRNU-based device identification.
Only recently, in \cite{Kirchner2019} sensor noise extraction has been improved by using a residual-based deep learning process that replaces the standard agnostic denoising procedure.

In this paper, we devise a data-driven approach for source identification based on PRNU and consider the scenario in which PRNU and images are geometrically synchronized. Differently from \cite{Kirchner2019}, our goal is to replace the PCE test in favour of an alternative method based on CNNs. 
Using deep learning to compare patches is a powerful approach \cite{Zagoruyko2015}.
It has been recently proposed for forensics applications 
in \cite{Huh2018, Mayer2020, Cozzolino2020} by means of Siamese networks, 
in order to learn distinctive data representations that highlight camera-based artifacts.
Unlike \cite{Zagoruyko2015}, our solution does not compare directly raw image pixels, but work on noise residuals.
In our scenario, patches to be compared are extracted from the estimated PRNU and the noise residual of the test image, which do not possess any type of periodicity and are hard to compress. 
For this reason, we avoid embedded representations and 
rely on the whole original information, by proposing a network architecture 
that learns the best similarity function for source identification.
Specifically, we design a 2-channel-based CNN, which, through proper training, is able to learn the best way to compare patches for device identification. 
We propose a shallow CNN architecture which turns out to be faster and more accurate than standard PCE approach. Furthermore, deeper CNNs are also investigated, enabling to achieve even better accuracy if processing time is not a constraint.

\section{Proposed method}
\label{sec:algorithm}

Given a set of images coming from device $d$, it is possible to estimate the device PRNU $\K_d$ through denoising and maximum-likelihood estimation \cite{Lukas2006, Chen2008}.
Given a test image $\I$, the noise residual $\W$ can be extracted using a denoising-based procedure \cite{Chen2008}.
Standard PRNU-based device attribution methods compare $\W$ and $\K_d$ in order to detect a possible match \cite{Chen2007}.
In this work, we propose to perform this comparison through a CNN.
For each pair of query image $\I$ and candidate device $d$, we feed $\K_d$ and $\W$ to the CNN, assuming they are geometrically synchronized. The network returns a CNN-based identification score $\cs$ which is directly associated to the coherence between the image and the device. 
Analyzing $\cs$ we can infer whether $\I$ belongs to $d$ or not.

\begin{figure}[t!]
	\centering
	\includegraphics[width=.9\columnwidth]{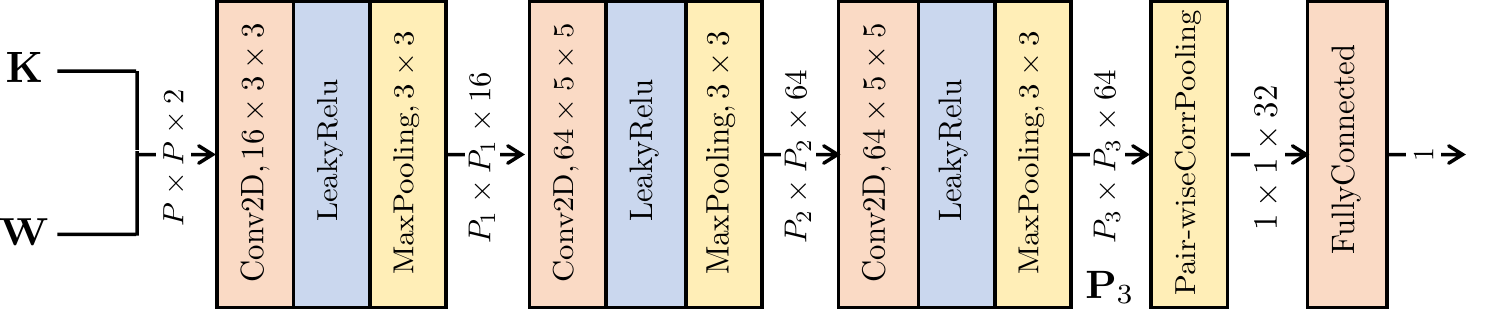}
	\caption{Scheme of the proposed pair-wise correlation network (PCN).}
	\label{fig:cnn_pcn3}
\end{figure}

\subsection{Network architectures}
As the proposed 2-channel-based approach is independent from the used CNN, we investigate different architectures.

With the specific purpose of testing a shallow network enabling fast computations, the first proposed architecture is drawn using only $ 3 $ convolutional layers as depicted in Fig.~\ref{fig:cnn_pcn3}.
Then, a pair-wise correlation pooling layer and a fully-connected layer follow to obtain a single score $\cs$.
The pair-wise correlation pooling layer is inspired to the conventional metrics used to identify the source device on images, i.e., normalized cross-correlation and $ \pce $.
Since these are based on cross-correlation between noise residual and PRNU, the pair-wise correlation pooling computes a correlation as well between pairs of input features.
Being $ \P_3 $ the input of the pair-wise correlation pooling, the output of the layer is defined as:
\begin{equation}
 [\P_4]_n = \frac{1}{P_3^2} \sum_{p_i=1}^{P_3}  \sum_{p_j=1}^{P_3} [\P_3]_{p_i, p_j, 2n-1} \cdot  [\P_3]_{p_i, p_j, 2n}, n \in [1, 32], 
\end{equation}
where $ n$ is the channel index.
The pair-wise correlation pooling layer can be seen as a simplified version of bilinear pooling layer \cite{Lin2015}. 
Actually, bilinear pooling computes the correlation between \textit{all} the features maps, whereas pair-wise pooling evaluate this only for \textit{adjacent} pairs of features.
The main motivation behind pair-wise pooling is to provide fast computations. Indeed, evaluating the complete cross-correlation between all feature maps may be time-consuming.

The other networks we propose are known in literature as Inception-ResNet V2 \cite{Szegedy2017} and EfficientNet models \cite{Tan2019}, which achieve very good results in computer vision tasks.
Differently from the first solution which is trained from scratch, we initialize the network weights of EfficientNet and Inception models
using the weights trained on Imagenet database \cite{imagenet2009}.

For brevity's sake, we hereafter investigate three EfficientNet models with increasing depth \cite{Tan2019}: B0, B2 and B4. 
We denote the networks as PCN (Pair-wise Correlation Net), EFFB0, EFFB2, EFFB4 (EfficientNet)  and INC (Inception).

\subsection{Network training}

\begin{figure}[t!]
	\centering
	\includegraphics[width=.9\columnwidth]{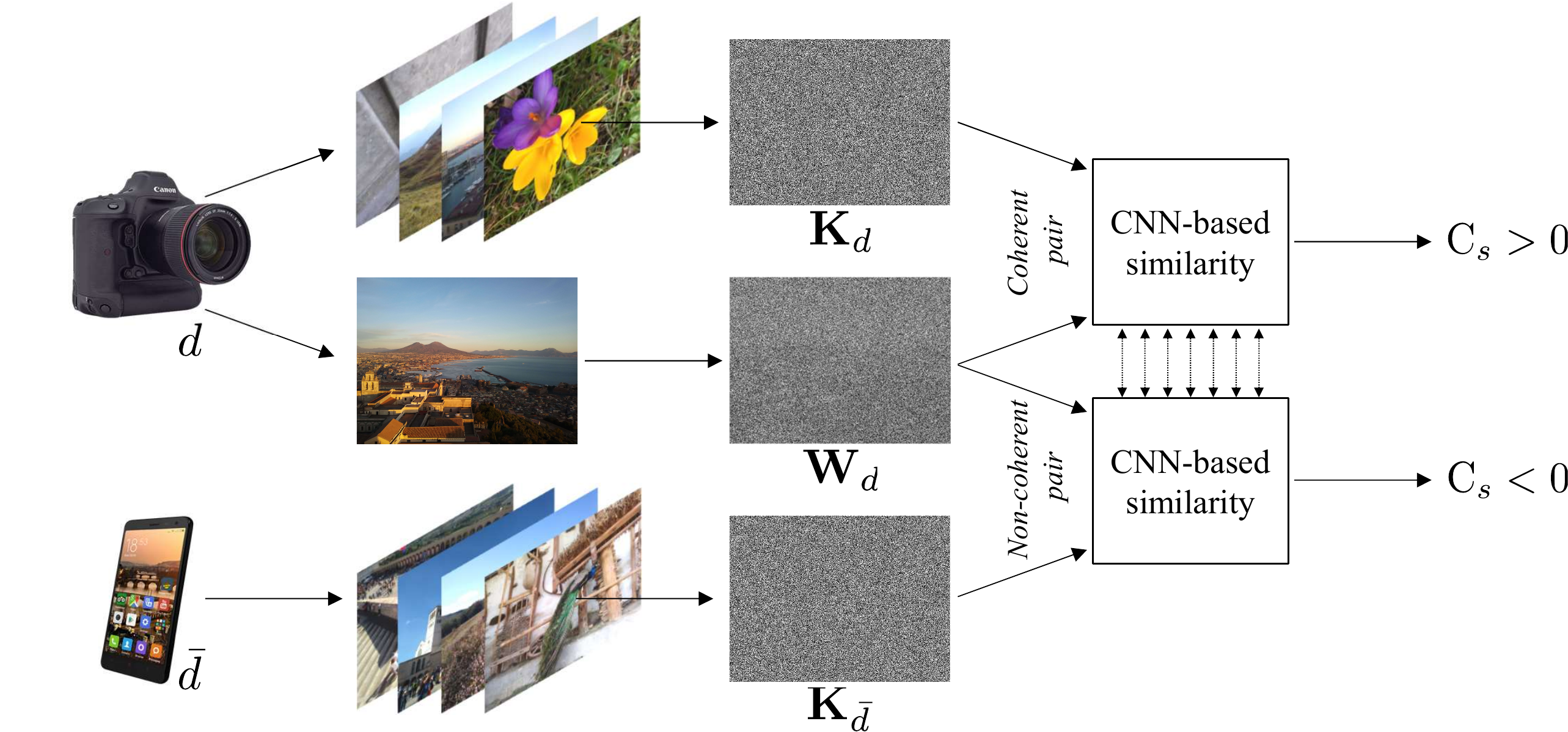}
	\caption{Sketch of the proposed training strategy. For every device $ d \in \Dtv$, the network is fed using noise residuals coming from device $d$, paired with PRNU $\K_d$ (coherent pair) and with PRNU $\K_{\bar{d}}$ coming from a different device (non-coherent pair).
	The CNN is able to learn a similarity measure $\mathrm{C}_s$ for the source identification task.}
	\label{fig:cnn_training}
\end{figure}

To train the network, we assume that a set $\Dtv$ of $N_d$ devices is available.
For each device, we assume having a PRNU estimate $\K_d$ (obtained from a set of flat-field images) as well as a set of natural images.
From each natural image belonging to device $d$, we extract the noise residue $\W_d$.

As shown in Fig.~\ref{fig:cnn_training}, the network is trained by concurrently feeding it with coherent pairs (i.e., PRNU and residue of the same device), and non-coherent pairs (i.e., PRNU and residue of diverse devices). 
In order to provide both coherent and non-coherent cases for each device inside one batch of data, we consider a batch-size of twice the cardinality of $\Dtv$.
For every $ d \in \Dtv$, the coherent pair $ (\K_d, \W_d) $ is created by randomly picking $  \W_d $ from the set of residues of device $d$, while the non-coherent pair $ (\K_{\bar{d}}, \W_d)$ takes $\K_{\bar{d}}$ from another device $\bar{d}$ different from $d$, $\bar{d} \in \Dtv$.

As far as the loss function is concerned, 
we face the problem as a binary classification problem, adopting the standard sigmoid cross-entropy loss.
In other words, label $1$ is assigned to coherent pairs and label $0$ otherwise.
The network learns to return a score $\cs$, which is high for coherent pairs and low for non-coherent ones.
We stop the training when the accuracy (i.e., the fraction of correct predictions) over the validation set is maximized. 
We use Adam optimization algorithm \cite{kingma2014} with learning rate and patience initialized at $ 0.001 $ and $ 30 $, respectively,
for a maximum number of $500$ epochs.

\subsection{Network deployment}

Once the CNN is trained, we can use it in two scenarios:
\begin{itemize}
	\item \textit{closed-set} scenario: given an image, 
	we aim at identifying the source among a finite pool of devices;
	\item \textit{open-set} scenario: given an image and a candidate device, we aim at inferring if the device shot that image or not.
\end{itemize}
To solve these problems, we always feed the network with pairs of PRNU and image residue.
Notice that, despite the CNN is trained over a specific set of devices $\Dtv$, it can be used also to compare PRNUs and residues from devices never seen during training (i.e., $d \notin \Dtv$).
As a matter of fact, the CNN learns how to compute a distance measure between a PRNU and a residue, thus it is not bounded to work on a closed-set of devices.

\section{Experimental analysis}
\label{sec:results}
In this section, we first describe the dataset, the experimental setup and the evaluation metrics, then we report numerical results discussing the main achievements.

\subsection{Dataset}
\label{subsec:sdi_cnn_dataset}
The input to the network is always a pair of PRNU and image noise residue.
These terms are assumed to be synchronized, i.e., no pixel misalignment is present.
In our experiments, we always crop the central $P \times P $ pixel region from each image in order to limit network complexity.
We let $P$ range from $P = 80$ to $P = 720$ to study the relationship between image resolution and accuracy.
As commonly done in CNN-based solutions, we normalize both PRNUs and residues by their standard deviation before feeding them to the network.

In order to test our method on a significant amount of devices, we consider both the Dresden image database \cite{Gloe2010} and the Vision dataset \cite{Shullani2017}.
To avoid excessively old camera models, we only pick devices whose imagery have resolution greater than $720 \times 720$ pixels.
For each device we exclusively investigate JPEG compressed images, as these represent the most frequent data for a forensics analyst to deal with.

Focusing on computing reliable PRNU fingerprints, we select devices if at least $ 25 $ JPEG images depicting scenes of flat surfaces are available.
We end up with a total amount of $87$ devices, $ 55 $ from Dresden and $32$ from Vision.
To estimate the PRNU we select only flat-field images,
whereas, 
in order to fairly test our method, we make use of images showing natural scenes taken from indoor and outdoor scenarios.
We work with more than $ 12000 $ images belonging to Dresden and almost $7000$ images from Vision. On average, there are more than $200$ natural images per device, randomly split in three disjoint datasets: training ($50\%$ of the images), validation ($25\%$) and evaluation (remaining $25\%$).

The training dataset $ \Dtv $ includes $ 20 $ devices from Vision and $16$ 
devices 
from Dresden.
It is worth mentioning that $\Dtv$ has been built using only
devices of 
\emph{different} models.
In doing so, we avoid introducing an important constraint into the learning process, that is, training using cameras of the same model.
Indeed, this may help enhancing the final performance, although requiring various instances of the same model at investigation side, which is actually unlikely to happen.
Results are computed on images from the evaluation set of all $87$ devices.
Thereby, we are simulating a real situation
in which 
the system is tested over \textit{unknown} camera models as well.

\subsection{Evaluation metrics}
\label{subsec:sdi_cnn_metrics}

In the closed-set scenario, we test the residue of the query image against all $87$ PRNUs. 
The camera returning the highest score $\cs$ is associated to the image.
To assess the attribution accuracy, we consider all query residues in the evaluation set for each device. 
The closed-set accuracy score $\Acs$ is defined as the average fraction of correct predictions per device.

The open-set scenario is evaluated as a binary classification problem: distinguishing correlating pairs of PRNU-residue from non-correlating ones. 
Notice that we do not threshold the CNN score $\cs$ associated to each test image, but we resort to receiver operating characteristic (ROC) curves. 
For each camera we consider all noise residues from that camera
as positive samples, whereas the set of negatives includes 
all the
residues not belonging to that camera.
Each ROC curve draws the relationship between true positive rate ($ \textrm{TPR} $) and false positive rate ($ \textrm{FPR} $), computed over the set of available devices. 
As compact metric we use $\Aos$, defined as the area under the curve ($ \auc $) for the open-set problem. 
The goal is to achieve a high value of $\Aos$, ideally 1.

\subsection{Results}
All tests have been run on a workstation equipped with one Intel\textsuperscript{\scriptsize{\textregistered}} Core\textsuperscript{\tiny{$ \mathsf{TM} $}} i9-9980XE (36 Cores @3.00 GHz), RAM 126 GB, one QUADRO P6000 (3840 CUDA Cores @1530 MHz), 24 GB, running Ubuntu 18.04.2.
For the sake of clarity, we report at this link\footnote{\url{https://github.com/polimi-ispl/cnn-fast-sdi}} the code used for the experiments, together with some additional materials to analyze results in details.

\begin{figure}[t]
	\centering
	\includegraphics[width=.9\columnwidth]{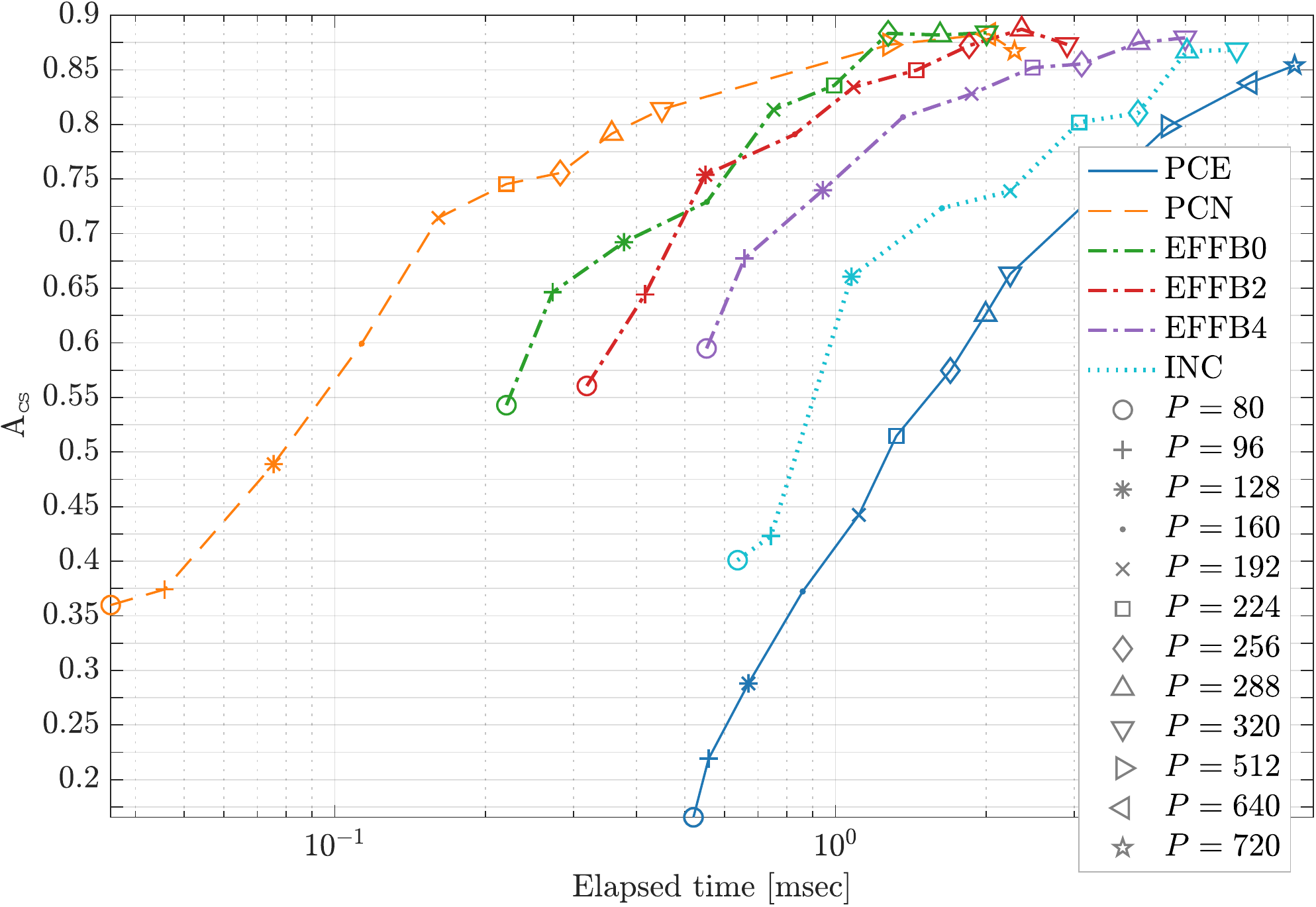}
	\caption{Accuracy $\Acs$ as a function of time [milliseconds] and crop size $P$.}
	\label{fig:sdi_cnn_acs}
\end{figure}

\vspace{.5em}\noindent\textbf{Closed-set scenario.}
Fig.~\ref{fig:sdi_cnn_acs} shows results for the closed-set scenario.
$\Acs$ is depicted as a function of the average computational time for testing one pair of PRNU-residue, and of the crop size $P$. 
For EFF and INC models we draw results only for $P \leq 320$, as larger image dimensions
would require additional GPU memory.
As state-of-the-art comparison, we perform the classic $ \pce $ test \cite{Chen2007} between each noise residual in the evaluation dataset and all available PRNUs.

All the proposed CNNs show better accuracy than PCE, which achieves similar results only for the highest $P$ values.
EFF models always outperform results of INC, which actually requires more computation time as well.
Whenever only very small patches are available (i.e., $P \leq 160 $), EFFB4 represents a very good solution to achieve $\Acs$ even larger than $80\%$. 
On the contrary, if time is a strong constraint, PCN is a good trade-off between accuracy and required computation time. 
For any patch-size, $ \pce $ and other CNNs require at least $4$ times as much the computations of PCN.

For what concerns the required computational times, it is worth noting that the main advantage with respect to $ \pce $ is the possibility of feeding the CNNs with multiple pairs of PRNU-residue. 
These pairs can be processed in parallel by the GPU, at least until there is available memory for storing data.
As a consequence, the larger the amount of candidate devices, the higher the temporal benefit compared to $ \pce$.
For instance, when testing a query image over $87$ candidate cameras, we always take up less than 8 GB of GPU memory even using the INC configuration.
Whether more devices were available, the average testing time would remain basically unchanged until reaching the maximum memory size.

\begin{figure}[t]
	\centering
	\includegraphics[width=.9\columnwidth]{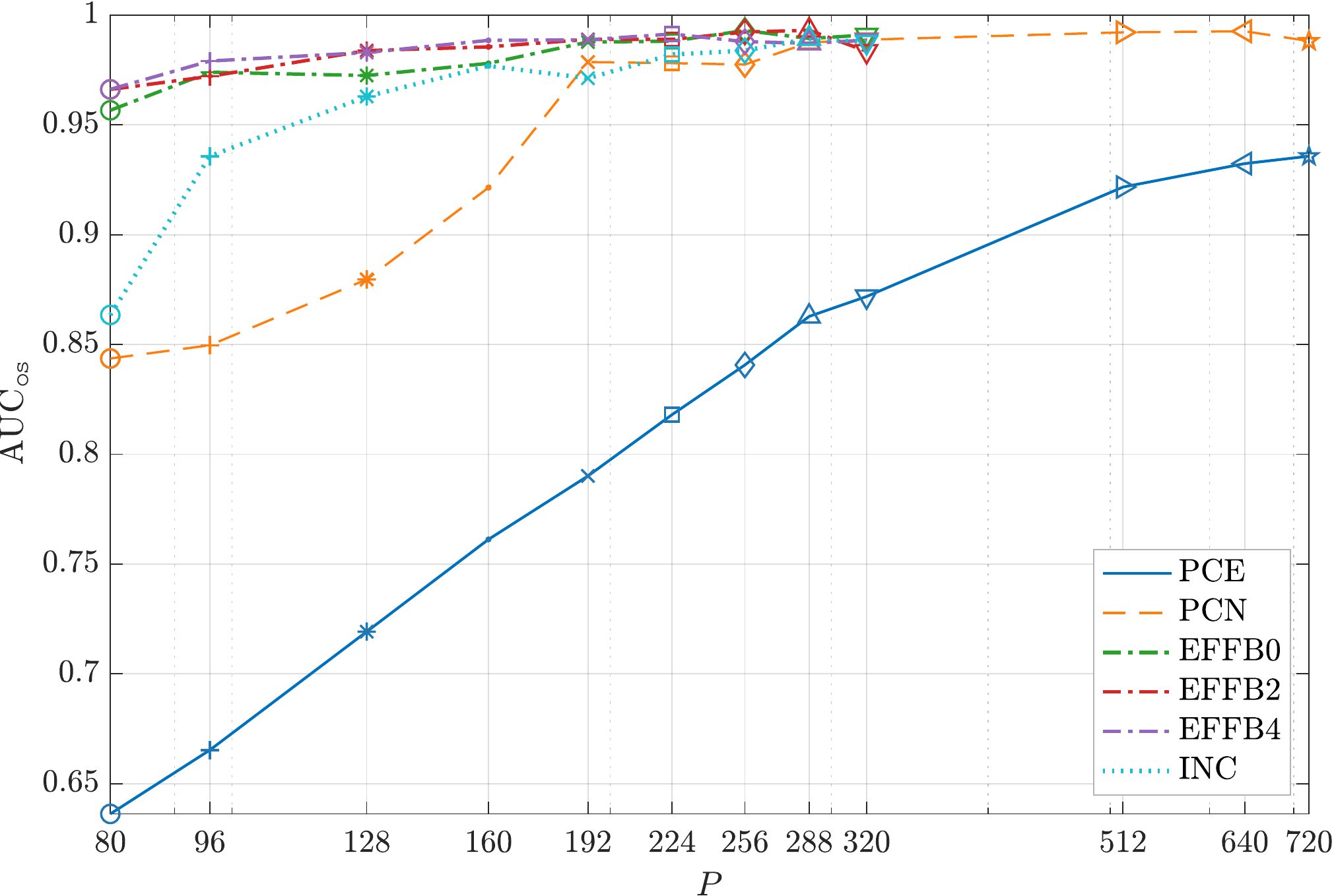}
	\caption{Area under the ROC curve $\Aos$ as a function of crop size $P$.}
	\label{fig:sdi_cnn_aos}
\end{figure}

\vspace{.5em}\noindent\textbf{Open-set scenario.}
Fig.~\ref{fig:sdi_cnn_aos} reports results for the open-set scenario. 
All proposed architectures outperform PCE for any $P$ value, even though the highest $\Aos$ is obtained by EFF models.
Notice that we are not analyzing $\Aos$ as a function of the required time, since the open-set scenario reduces investigations to only one pair of PRNU-residue.
In this scenario, CNNs and $\pce$ report comparable computational time, in the order of few milliseconds.
However, as previously shown, whenever various images should be tested against the same PRNU, we could exploit the data parallelization property of GPU to test multiple images together, thus saving important computation time.

\vspace{.5em}\noindent\textbf{JPEG re-compression.}
In order to simulate scenarios where images underwent some post-processing operations, we test our method on double JPEG-compressed images.
This step is purposely designed to simulate real case setups, in which only images from social networks (typically undergoing at least two compression steps) may be available.
In this vein, we re-compress all the images using JPEG compression with quality factors $80$ and $90$.
Then, we extract the noise residuals \cite{Chen2008}, defining them as $\W_{\textrm{JPEG}}$.
As far as PRNUs are concerned, we can distinguish two scenarios: (i) whether flat-field single-compressed images are available, exploit the PRNUs $\K$ estimated from these; (ii) if only flat-field double-compressed images are available, use them to compute the PRNUs defined as $\K_{\textrm{JPEG}}$.
As the analyst might have only double-compressed test data, it is important to understand whether she/he can exploit the network trained on single-compressed data, or some sort of domain adaptation must be applied.
To this purpose, we evaluate different setups, considering training and testing on single and/or double-compressed data.

\begin{figure}[t]
	\centering
	\includegraphics[width=.9\columnwidth]{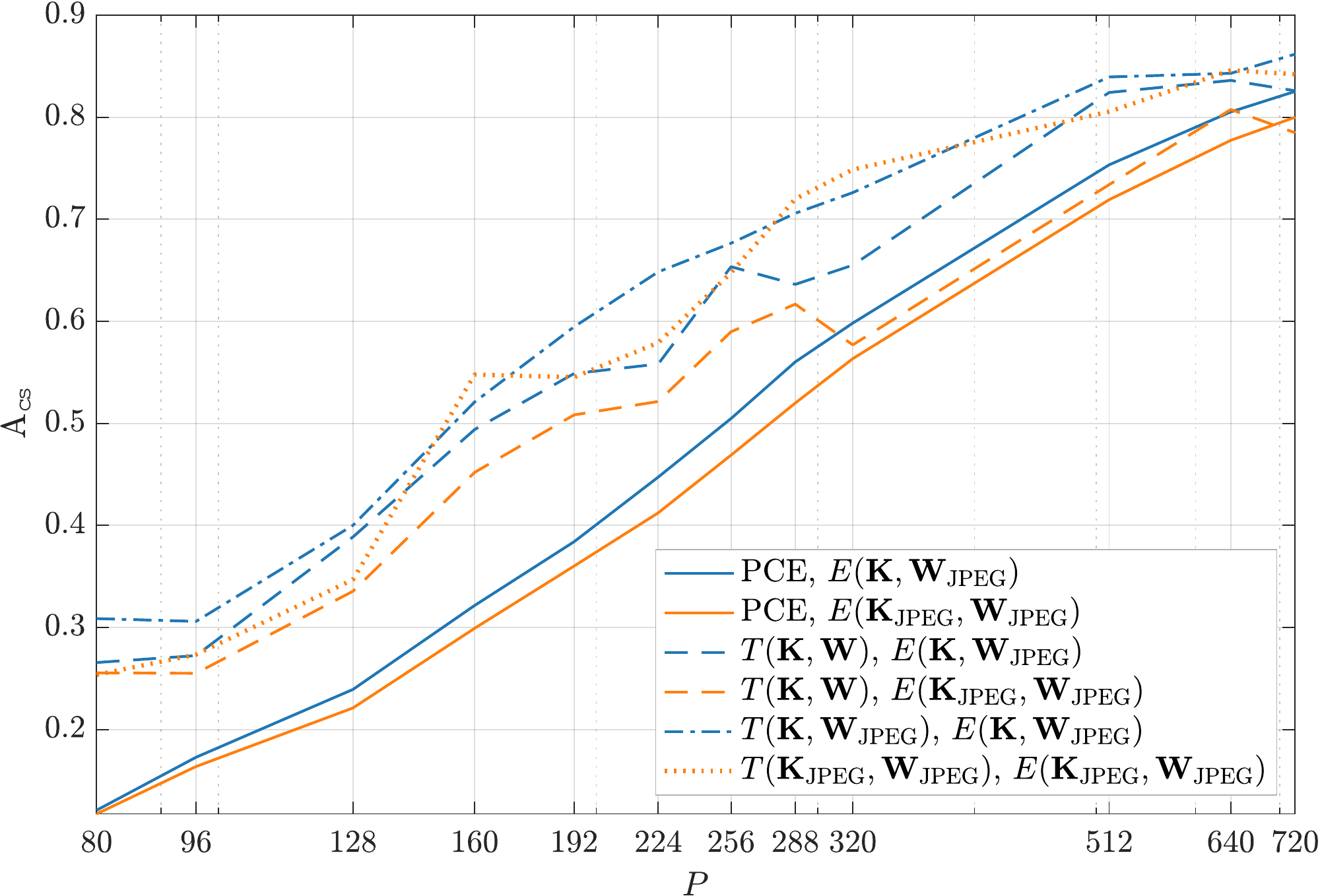}
	\caption{Accuracy $\Acs$ as a function of crop size $P$, considering double-JPEG compressed dataset with quality 90 and PCN network. $T(\cdot)$ and $E(\cdot)$ refer to training and evaluation datasets, respectively.}
	\label{fig:accuracy_jpeg}
\end{figure}

Experiments show that we can achieve also $20$ percentage points of accuracy more than PCE for both JPEG qualities $80$ and $90$. 
This 
improvement is evident especially for the smallest patch-sizes (i.e., $P < 320$).
For instance, Fig.~\ref{fig:accuracy_jpeg} depicts the results achieved on the closed-set scenario using PCN configuration on double-compressed dataset with JPEG quality $90$.
As expected, it is worth noting that 
performing domain adaptation (i.e., training over the re-compressed dataset)
always helps enhancing the performance with respect to training on the single-compressed dataset, even though every reported curve outperforms the corresponding PCE results. 

\section{Conclusions}
\label{sec:conclusions}

In this paper we propose a fast solution to the image device identification problem. In particular, we leverage PRNU and standard image noise residual extraction, but we substitute the correlation stage with a 2-channel-based CNN,
able to learn the best similarity function for source device identification.
Our method proves to be faster than PCE in case a large amount of potential provenance 
devices is investigated,
requiring much less query image content to obtain enhanced attribution accuracy, given that images and PRNUs are assumed to be geometrically synchronized.
Eventually, we evaluate the proposed methodology on images
undergone to double-JPEG compression, in order to simulate more complex scenarios.

Given these promising results, our future work will focus on
including the noise residual extraction step into the learning process as suggested by \cite{Kirchner2019}, as well as investigating how to deal with potential pixel misalignment between the image and the fingerprint.
Moreover, video sequences will be the natural continuation of our research.
Indeed, compression artifacts \cite{Chuang2011, Altinisik2019} and stabilization issues \cite{Iuliani2019, Mandelli2019} make video source identification a complex task of undoubted interest.

\ifCLASSOPTIONcaptionsoff
  \newpage
\fi

\bibliographystyle{IEEEtran}
\bibliography{bibliography}

\begin{thebibliography}{10}
\providecommand{\url}[1]{#1}
\csname url@samestyle\endcsname
\providecommand{\newblock}{\relax}
\providecommand{\bibinfo}[2]{#2}
\providecommand{\BIBentrySTDinterwordspacing}{\spaceskip=0pt\relax}
\providecommand{\BIBentryALTinterwordstretchfactor}{4}
\providecommand{\BIBentryALTinterwordspacing}{\spaceskip=\fontdimen2\font plus
\BIBentryALTinterwordstretchfactor\fontdimen3\font minus
  \fontdimen4\font\relax}
\providecommand{\BIBforeignlanguage}[2]{{%
\expandafter\ifx\csname l@#1\endcsname\relax
\typeout{** WARNING: IEEEtran.bst: No hyphenation pattern has been}%
\typeout{** loaded for the language `#1'. Using the pattern for}%
\typeout{** the default language instead.}%
\else
\language=\csname l@#1\endcsname
\fi
#2}}
\providecommand{\BIBdecl}{\relax}
\BIBdecl

\bibitem{Lukas2006}
J.~Lukas, J.~Fridrich, and M.~Goljan, ``Digital camera identification from
  sensor pattern noise,'' \emph{IEEE Transactions on Information Forensics and
  Security}, vol.~1, no.~2, pp. 205--214, 2006.

\bibitem{Chen2008}
M.~Chen, J.~Fridrich, M.~Goljan, and J.~Luk{\'a}s, ``Determining image origin
  and integrity using sensor noise,'' \emph{IEEE Transactions on Information
  Forensics and Security}, vol.~3, no.~1, pp. 74--90, 2008.

\bibitem{Chierchia2014}
G.~Chierchia, G.~Poggi, C.~Sansone, and L.~Verdoliva, ``{A Bayesian-MRF
  approach for PRNU-based image forgery detection},'' \emph{IEEE Transactions
  on Information Forensics and Security}, vol.~9, no.~4, pp. 554--567, 2014.

\bibitem{Chen2007}
M.~Chen, J.~Fridrich, M.~Goljan, and J.~Luk{\'a}{\v{s}}, ``Source digital
  camcorder identification using sensor photo response non-uniformity,'' in
  \emph{Security, Steganography, and Watermarking of Multimedia Contents IX},
  vol. 6505, 2007.

\bibitem{Lin2017}
X.~Lin and C.~Li, ``{Large-scale image clustering based on camera
  fingerprints},'' \emph{IEEE Transactions on Information Forensics and
  Security}, vol.~12, no.~4, pp. 793--808, 2017.

\bibitem{Marra2017}
F.~Marra, G.~Poggi, C.~Sansone, and L.~Verdoliva, ``{Blind PRNU-based image
  clustering for source identification},'' \emph{IEEE Transactions on
  Information Forensics and Security}, vol.~12, no.~9, pp. 2197--2211, 2017.

\bibitem{Phan2018}
Q.~Phan, G.~Boato, and F.~{De Natale}, ``{Accurate and scalable image
  clustering based on sparse representation of camera fingerprint},''
  \emph{IEEE Transactions on Information Forensics and Security}, vol.~14,
  no.~7, pp. 1902--1916, 2018.

\bibitem{Goljan2010}
M.~Goljan and J.~Fridrich, ``Managing a large database of camera
  fingerprints,'' in \emph{Media Forensics and Security II, SPIE Electronic
  Imaging Symposium}, 2010.

\bibitem{Bayram2012}
S.~Bayram, H.~Sencar, and N.~Memon, ``{Efficient Sensor Fingerprint Matching
  Through Fingerprint Binarization},'' \emph{IEEE Transactions on Information
  Forensics and Security}, vol.~4, no.~4, pp. 1404--1413, 2012.

\bibitem{Bayram2015}
------, ``{Sensor fingerprint identification through composite fingerprints and
  group testing},'' \emph{IEEE Transactions on Information Forensics and
  Security}, vol.~10, no.~3, pp. 597--612, 2015.

\bibitem{Valsesia2015}
D.~Valsesia, G.~Coluccia, T.~Bianchi, and E.~Magli, ``{Compressed fingerprint
  matching and camera identification via random projections},'' \emph{IEEE
  Transactions on Information Forensics and Security}, vol.~10, no.~7, pp.
  1472--1485, 2015.

\bibitem{Bondi2019}
L.~Bondi, P.~Bestagini, F.~P{\'e}rez-Gonz{\'a}lez, and S.~Tubaro, ``{Improving
  PRNU compression through preprocessing, quantization, and coding},''
  \emph{IEEE Transactions on Information Forensics and Security}, vol.~14,
  no.~3, pp. 608--620, 2019.

\bibitem{Taspinar2017}
S.~Taspinar, H.~Sencar, S.~Bayram, and N.~Memon, ``Fast camera fingerprint
  matching in very large databases,'' in \emph{IEEE International Conference on
  Image Processing}, 2017.

\bibitem{Tuama2016}
A.~Tuama, F.~Comby, and M.~Chaumont, ``Camera model identification with the use
  of deep convolutional neural networks,'' in \emph{IEEE International Workshop
  on Information Forensics and Security}, 2016.

\bibitem{Bondi2017}
L.~Bondi, L.~Baroffio, D.~Guera, P.~Bestagini, E.~Delp, and S.~Tubaro, ``First
  steps toward camera model identification with convolutional neural
  networks,'' \emph{IEEE Signal Processing Letters}, vol.~24, no.~3, pp.
  259--263, 2017.

\bibitem{Caldelli2018}
R.~Caldelli, I.~Amerini, and C.~Li, ``{PRNU-based Image Classification of
  Origin Social Network with CNN},'' in \emph{European Signal Processing
  Conference}, 2018, pp. 1357--1361.

\bibitem{Bonettini2018}
N.~Bonettini, L.~Bondi, D.~G{\"u}era, S.~Mandelli, P.~Bestagini, S.~Tubaro, and
  E.~J. Delp, ``{Fooling PRNU-based detectors through convolutional neural
  networks},'' in \emph{European Signal Processing Conference (EUSIPCO)}, 2018,
  pp. 957--961.

\bibitem{Kirchner2019}
M.~Kirchner and C.~Johnson, ``{SPN-CNN: Boosting Sensor-Based Source Camera
  Attribution with Deep Learning},'' in \emph{IEEE International Workshop on
  Information Forensics and Security}, December 2019.

\bibitem{Zagoruyko2015}
S.~Zagoruyko and N.~Komodakis, ``Learning to compare image patches via
  convolutional neural networks,'' in \emph{IEEE Conference on Computer Vision
  and Pattern Recognition}, 2015.

\bibitem{Huh2018}
M.~Huh, A.~Liu, A.~Owens, and A.~Efros, ``Fighting fake news: Image splice
  detection via learned self-consistency,'' in \emph{European Conference on
  Computer Vision}, 2018.

\bibitem{Mayer2020}
O.~Mayer and M.~Stamm, ``{Forensic Similarity for Digital Images},'' \emph{IEEE
  Transactions on Information Forensics and Security}, vol.~15, pp. 1331--1346,
  2020.

\bibitem{Cozzolino2020}
D.~{Cozzolino} and L.~{Verdoliva}, ``Noiseprint: A cnn-based camera model
  fingerprint,'' \emph{IEEE Transactions on Information Forensics and
  Security}, vol.~15, pp. 144--159, 2020.

\bibitem{Lin2015}
T.-Y. Lin, A.~RoyChowdhury, and S.~Maji, ``Bilinear cnn models for fine-grained
  visual recognition,'' in \emph{Proceedings of the IEEE international
  conference on computer vision}, 2015, pp. 1449--1457.

\bibitem{Szegedy2017}
C.~Szegedy, S.~Ioffe, V.~Vanhoucke, and A.~A. Alemi, ``Inception-v4,
  inception-resnet and the impact of residual connections on learning,'' in
  \emph{Thirty-first AAAI conference on artificial intelligence}, 2017.

\bibitem{Tan2019}
M.~Tan and Q.~V. Le, ``Efficientnet: Rethinking model scaling for convolutional
  neural networks,'' \emph{arXiv preprint arXiv:1905.11946}, 2019.

\bibitem{imagenet2009}
J.~Deng, W.~Dong, R.~Socher, L.-J. Li, K.~Li, and L.~Fei-Fei, ``{ImageNet: A
  Large-Scale Hierarchical Image Database},'' in \emph{IEEE Conference on
  Computer Vision and Pattern Recognition}, 2009.

\bibitem{kingma2014}
D.~P. Kingma and J.~Ba, ``Adam: A method for stochastic optimization,''
  \emph{arXiv preprint arXiv:1412.6980}, 2014.

\bibitem{Gloe2010}
T.~Gloe and R.~B{\"o}hme, ``{The 'Dresden Image Database' for benchmarking
  digital image forensics},'' \emph{Journal of Digital Forensic Practice},
  vol.~3, pp. 150--159, 2010.

\bibitem{Shullani2017}
D.~Shullani, M.~Fontani, M.~Iuliani, O.~Al~Shaya, and A.~Piva, ``{VISION: a
  video and image dataset for source identification},'' \emph{EURASIP Journal
  on Information Security}, vol. 2017, no.~1, p.~15, 2017.

\bibitem{Chuang2011}
W.-H. Chuang, H.~Su, and M.~Wu, ``Exploring compression effects for improved
  source camera identification using strongly compressed video,'' in \emph{IEEE
  International Conference on Image Processing}, 2011, pp. 1953--1956.

\bibitem{Altinisik2019}
E.~Altinisik, K.~Tasdemir, and H.~Sencar, ``{Mitigation of H.264 and H.265
  video compression for reliable PRNU estimation},'' \emph{IEEE Transactions on
  Information Forensics and Security}, vol.~15, pp. 1557--1571, 2019.

\bibitem{Iuliani2019}
M.~Iuliani, M.~Fontani, D.~Shullani, and A.~Piva, ``Hybrid reference-based
  video source identification,'' \emph{Sensors}, vol.~19, no.~3, p. 649, 2019.

\bibitem{Mandelli2019}
S.~{Mandelli}, P.~{Bestagini}, L.~{Verdoliva}, and S.~{Tubaro}, ``Facing device
  attribution problem for stabilized video sequences,'' \emph{IEEE Transactions
  on Information Forensics and Security}, vol.~15, pp. 14--27, 2020.

\end{thebibliography}

\end{document}